\documentclass[10pt,journal,compsoc]{IEEEtran}

\ifCLASSOPTIONcompsoc
  \usepackage[nocompress]{cite}
\else
  \usepackage{cite}
\fi

\newcommand\citep{\cite}
\usepackage{amsmath}
\usepackage{amssymb}
\usepackage{mathtools} 
\usepackage{booktabs}
\usepackage{multirow}
\usepackage{comment}
\usepackage{enumitem, array}
\usepackage{multicol}
\usepackage[author={Henglin Shi}]{pdfcomment}

\ifCLASSINFOpdf
  
\else
  
\fi

\newcommand\MyBox[2]{
  \fbox{
    \lower0.75cm
    \vbox to 1.7cm{
        \vfil
        \hbox to 1.7cm{
            \hfil\parbox{1.4cm}
            {
            \begin{center}
                #1#2
            \end{center}
            }
            \hfil
        }
        \vfil
    }%
  }%
}

\usepackage[caption=false,font=footnotesize]{subfig}
\usepackage{stfloats}
\usepackage{array}
\usepackage{graphicx}
\usepackage{multirow}
\usepackage[rgb]{xcolor}
\usepackage{threeparttable}
\begin{document}

\title{iMiGUE-3K: A Large-Scale Benchmark for Micro-Gesture Analysis with Self-Supervised Learning}


\author{Chengyan~Wang$^1$, Haoyu~Chen$^1$, Hui~Wei$^1$, Yueyi~Yang$^1$, Yunquan Chen$^2$, and~Guoying~Zhao$^{1,3*}$
\thanks{$^{1}$Chengyan~Wang, Haoyu~Chen, Hui~Wei, Yueyi~Yang and Guoying~Zhao are with CMVS, University of Oulu, Finland. $^{2}$ Yunquan Chen is with KTH Royal Institute of Technology, Sweden. $^{3}$Guoying~Zhao is also with the ELLIS Institute Finland.}
\thanks{$^{*}$Corresponding author emails: guoying.zhao@oulu.fi}

}


\IEEEtitleabstractindextext{%

\begin{abstract}
Emotion understanding is a fundamental challenge in affective computing and artificial intelligence. While existing approaches predominantly focus on facial expressions and speech, they often overlook the rich emotional cues conveyed through body language. Recently, micro-gestures (MGs)—unintentional, subconscious movements driven by inner feelings—have attracted increasing attention as an alternative to other cues. MGs have its unique advantages, e.g.\, they offer a privacy-preserving solution, as they do not rely on sensitive biometric data. However, there are no existing large-scale datasets supporting the pre-training of the MG foundation model.
To advance MG research, we present a new benchmark for micro-gesture-based emotion understanding, featuring key contributions with a novel dataset (\textbf{iMiGUE-3K}) and a series of foundation models for different tasks. Using a model-based crowd-sourcing data collection strategy, we construct iMiGUE-3K—the largest MG dataset to date. It comprises video recordings from 332 distinct professional tennis players' public press appearances over the past seven years, totaling more than 3.4K long video clips and 37 million frames. The dataset includes 32 micro-gesture classes with rich descriptive annotations, making it the first large-scale, in-the-wild, video dataset for fine-grained gesture-based emotion analysis. To efficiently train on iMiGUE-3K, we propose a novel data sampling strategy to obtain high-quality samples.
Built on iMiGUE-3K, we propose \textbf{MG-FMs}, a discriminative foundation model for transferable gesture presentation learning. Based on the foundation model, we establish five comprehensive evaluation tasks: MG recognition (unsupervised, semi-supervised, supervised), MG retrieval,
MG emotion recognition. Our systematic evaluation of representative methods demonstrates that micro-gesture-based analysis significantly improves emotion understanding.
We hope this work can provide comprehensive tools for MG analysis and set a solid foundation for future research in psychological diagnostics, affective computing, and advanced human-computer interaction. Source code and dataset are available.

\end{abstract}

\begin{IEEEkeywords}
Micro-gesture, gesture recognition, emotion analysis, affective computing, large language models, large vision models, multi-modal learning
\end{IEEEkeywords}}

\maketitle

\IEEEdisplaynontitleabstractindextext

\IEEEpeerreviewmaketitle

\IEEEraisesectionheading{\section{Introduction}\label{sec:introduction}}

\IEEEPARstart{E}{motion} understanding is an essential capability for artificial intelligence (AI) technologies, as it is ever-present, influencing virtually all aspects of human activity, including thought processes and decision-making. According to psychological studies, body language is an essential clue to understanding human emotions. We, human beings, are arguably innately prepared to comprehend and respond to others’ emotional expressions via thousands of nonverbal behaviors including facial expressions, eye movements or gaze, tone of voices, gestures, touches, and the use of space. Automatic body language-based emotion understanding has attracted extensive interdisciplinary attention from many fields such as computer vision, psychology, with a considerable number of datasets have been proposed, e.g., posed facial expressions \cite{gross2010multipie, kanade2000facialdb, pantic2005facialdb, valstar2010mmi, yin2006facial3d, zhang2013highres3d}, spontaneous facial behaviors \cite{aung2015automatic, bartlett2006facial, dhall2017emotiw, lucey2011unbc, kollias2019affwild, mckeown2011semaine}, micro-expressions \cite{yan2014casme, li2013spontaneous, davison2016samm}, voice/speech \cite{schuller2011avec, schuller2012avec, mckeown2011semaine, ringeval2013recola}, social signals \cite{joo2015panoptic, joo2019socialai}, and multi-modal datasets with facial expressions and physiological signals \cite{soleymani2011multimodal,koelstra2011deap,mckeown2011semaine, ringeval2013recola}. 

\begin{figure}[ht!]
\centering
\includegraphics[width=8cm]{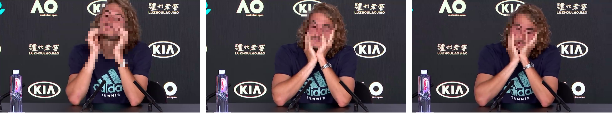}
\caption{We introduce iMiGUE-3K, an in-the-wild dataset of micro-gesture-based human emotion understanding. It's collected from post-match press conference videos of professional tennis players with annotations of micro-gestures, such as “cover face”, “fold arms”, and “cross fingers”. The iMiGUE aims to study: could machine 1) accurately recognize micro-gestures in the interview, 2) reasonably interpret those micro-gestures and 3) understand the emotional states of the player in a holistic way by identifying if the player has won or lost the match (positive or negative emotional states)}
\vspace{-0.5cm}
\label{fig1}
\end{figure}

Although advancements in computer vision and machine learning have made automatic emotion analysis more accessible with computational
methodologies and datasets, there are still significant gaps between current studies and the needs of real applications: 1) \textbf{Intentional and unintentional behavioral gestures.} Previous studies predominantly focused on illustrative or iconic gestures which are intentionally performed to convey meanings or emotions during interactions. However, people often suppress or hide their emotions, particularly negative ones. Research has identified a special group of gestures—unintentional behaviors elicited by inner feelings, such as rubbing hands due to stress—that provide valuable insights into suppressed emotions. Yet, machine learning efforts to understand these unintentional behaviors remain limited. 2) \textbf{Gap between behavior recognition and emotion understanding.} Existing datasets primarily aim to evaluate approaches for detecting and recognizing behavior prototypes, including gestures. In fact, the actual need of emotion AI is not merely to recognize certain behaviors, but
to uncover the emotion underneath. Research on interpreting these behaviors into accurate emotional states remains few. 3) \textbf{Sensitive biometric data concerns.} Many existing datasets involve sensitive biometric data, which plays a critical role in identity recognition across various applications. Despite its utility, biometric data is particularly vulnerable to identity theft, misuse, and unauthorized tracking. As privacy concerns escalate, greater emphasis must be placed on protecting individuals' biometric information.

Meanwhile, psychological studies \cite{ekman2009telling} showed that there are over 215 behaviors associated with psychological discomfort and most of them are not in the face. Specifically, previous studies \cite{aviezer2012body, axtell1991gestures, burgoon1989nonverbal} showed that there is a special group of gestures, the \textbf{Micro-Gestures (MGs)}, which are helpful to understand such suppressed or hidden emotions. Unlike deliberate gestures that are often used to convey specific meanings or facilitate communication—such as illustrative or iconic gestures that explicitly express emotions or attitudes \cite{illustrative}—MGs occur spontaneously or involuntarily in response to certain stimuli, particularly negative ones, e.g., rubbing hands due to stress, and the function of MGs is for relieving or protecting oneself from negative feelings rather than presenting for others. MGs have the potential to reveal person's true emotional state, as they are less likely to be consciously controlled or masked \cite{ekman2009telling}. Thus, being able to automatically recognize MGs would allow emotion understanding at a better level. Consider a post-match interview scenario, a player is interviewed by reporters over several question \& answer rounds (see Fig. \ref{fig1}). Some MGs could be observed, e.g., cross arms (defensive) and cover face (upset or ashamed), but it is hoped that the machine can understand (identify) if the player has a positive or negative feeling (e.g., caused by winning or losing of the match). It would be of great value if we can develop computer vision methods to capture and recognize these neglected clues for better emotion understanding.

Although several datasets for gesture-based emotion understanding have emerged, most of them remain small-scale. There is no large-scale data supporting foundation models in MG analysis. Consequently, models trained on these datasets are prone to overfitting and often show limited transferability. Motivated by the aforementioned observations and limitations, in this paper, we present the iMiGUE-3K dataset, the first large-scale MG dataset specifically designed for micro-gesture analysis with a comprehensive study of MGs from the machine learning aspect. The key contributions are summed up as follows:

1. \textbf{iMiGUE-3K: a large-scale micro-gesture dataset.} We introduce the first large-scale, in-the-wild MG dataset supporting the pretraining of MG foundation models, specifically collected from post-match tennis press conferences. This dataset contains 3,484 videos with over 17,800 manually labeled MG instances and 120K unlabeled clips for self-supervised learning (SSL). Each video is annotated with two levels: gesture categories and emotional states (win/loss). In total, this dataset covers 332 subjects, more than 3K videos, nearly 20K manually annotated MG instances, and 120K high-quality sampled clips for SSL, thereby enabling robust and generalizable learning in real-world settings.

2. \textbf{A broader benchmark for MG analysis.}
Beyond closed-set MG classification, we broaden the benchmark to include MG retrieval and MG-based emotion recognition, moving from category-level recognition toward instance-level retrieval and video-level affective analysis.

3. \textbf{Data Sampling Strategy for SSL}. To exploit large-scale unlabeled videos without exhaustive temporal annotation, we propose a sampling strategy that generates MG-length candidate clips using an empirical duration prior, facial-keypoint visibility constraints, and temporal non-overlap filtering. This strategy provides a scalable pretraining corpus for MG representation learning.

4. \textbf{MG-FMs: a series of transferable foundation models for micro-gesture representation.} We propose MG-FMs, a series of self-supervised pre-trained models that learn transferable representations for MGs. 

5. \textbf{Evaluation on multiple downstream tasks.} We conduct extensive experiments on linear evaluation, semi-supervised MG recognition, supervised MG recognition, MG retrieval, and MG-based emotion recognition. The results demonstrate the effectiveness of large-scale self-supervised pretraining for MG analysis. Notably, linear probing with the pretrained skeleton encoder achieves performance close to fully supervised skeleton baselines, and fine-tuning with the pretrained RGB-based model outperforms fully supervised methods, both of which indicate strong transferability across diverse MG categories.

The remainder of this paper is organized as follows. Sec.~\ref{sec2} reviews related work on emotion recognition and gesture analysis. Sec.~\ref{sec3} introduces the proposed iMiGUE-3K dataset, including its collection, annotation, and statistical properties. Sec.~\ref{sec4} presents the MG representation learning framework (MG-FMs). Sec.~\ref{sec:unsupervised_mg_recognition}, Sec.~\ref{sec:semisupervised_mg_recognition}, and Sec.~\ref{sec:supervised_mg_recognition} demonstrate benchmarking results of MG-FMs on gesture recognition. Sec.~\ref{sec:mg_retrieval} presents the results of MG retrieval on the proposed dataset. Sec.~\ref{sec6} illustrates the downstream task: emotion recognition. Finally, the conclusion is given in Sec.~\ref{sec8}.



\begin{table*}[htp]
\centering
\caption{The attributes comparison of iMiGUE-3K with other widely used datasets for recognizing gesture-based emotions. F/M:
Female/Male, C: Controlled (in-the-lab), U: Uncontrolled (in-the-wild), SP: Spontaneous, F: Face, G: Gesture, V: Voice.\label{table:datasetsumm}} 
\scriptsize
\resizebox{\linewidth}{!}{
\begin{tabular}{|l|>{\centering}p{0.65cm}  |  >{\centering}p{0.7cm}  |  >{\centering}p{1.8cm} |  >{\centering}p{1.2cm}  |  >{\centering}p{1cm}  |  c  |   >{\centering}p{0.5cm}    |  >{\centering}p{0.7cm}  |  c  |  >{\centering}p{1.1cm}|   >{\centering\arraybackslash}p{0.9cm}|}
\hline
Dataset       & \#Gest-ures & \#Emot-ions & \#Subjects (M/F) & \# Labeled Samples & \#Videos & Duration     & Con-text & Expre-ssion & Resolution & Modalities & Recogn -ition \\ \hline
FABO          & -           & 10          & 23 (12/11)        & 206        & 23        & 6 Min        & C       & Posed      & $1024\times768$ & F+G      & Isolated    \\ \hline
HUMANIE       & 8           & 8           & 10 (4/6)          & 240        & 240       & 5-180 Sec    & C       & Posed      & -          & F+G      & Isolated    \\ \hline
GEMEP         & -           & 18          & 10 (4/6)          & 7,000+      & 1,260      & -            & C       & Posed      & $720 \times 576$  & F+G      & Isolated    \\ \hline
THEATER       & -           & 8           & 10 (5/5)          & 258        & -         & -            & U       & SP         & -          & G          & Isolated    \\ \hline
Emilya        & 7           & 8           & -                 & 7,084       & 23        & 5.5 Sec      & C       & Posed      & $1280 \times 720$ & G          & Isolated    \\ \hline
LIRIS-ACCEDE  & 6           & 6           & 11 (6/5)          & -          & -         & 1 Min        & C       & Posed      & -          & F+G      & Isolated    \\ \hline
emoFBVP       & 23          & 23          & 64 (32/32)        & 1,380       & -         & 20-66 Sec    & C       & Posed      & $640 \times 480$  & F+G+V  & Isolated    \\ \hline
BoLD          & -           & 26          & 10 (-)            & 13,239      & 9,876      & -            & U       & SP         & -          & G          & Isolated    \\ \hline
SMG           &   17          &    2         & 40 (27/13)                 &       3,712     &   414        &       $\sim1$ Min       &      C   &       SP     &      $1920\times1080$      &    F+G        &    Holistic         \\ \hline
\textbf{iMiGUE-3K (ours)} & 32          & 2           & 332 (161/171)         &   ~17,800         & 3,484       & 0.2-41.3 Min & U       & SP         &    $1280\times720$        & F+G        & Holistic            \\ \hline
\end{tabular}
}
\label{tab1}
\end{table*}

\section{Related Work}
\label{sec2}

A person’s emotional state is often conveyed through
bodily expression. As such, analyzing body based activities, including action, gesture and posture are the popular
research topics \cite{kanade2000facialdb, pantic2005facialdb, yin2006facial3d, valstar2010mmi, gross2010multipie, zhang2013highres3d} in the community. However, these datasets focused on recognizing human activities (e.g., daily actions including jumping, clapping, etc.), rarely related to the emotional states. We limit our review on the related emotional gesture-based benchmarks. Then we review related work of gesture/action recognition.

\subsection{Emotional gesture-based datasets}

Gesture is one of the key cues of social communication, encompassing movements of the hands, head, and other body parts to express various emotions, thoughts, and intentions \cite{noroozi2018survey}. Table \ref{tab1} summarizes the attributes of widely used databases of emotional gestures. Early studies in this area focused mainly on acted or posed gestures. The \textbf{Tilburg University Stimulus set} \cite{schindler2008recognizing} collected photographic still images of actors enacting different emotions, though this dataset is excluded from Table \ref{tab1} due to its use of still images. The \textbf{FABO} database \cite{gunes2006bimodal} was one of the pioneering works, proposing video clips of posed prototype gestures for recognizing emotions, which were labeled with six basic emotions and four additional states: neutral, anxiety, boredom, and uncertainty.

Following these controlled, posed behaviors, researchers expanded emotional gesture analysis into various directions. For instance, in the HUMAINE project \cite{castellano2007recognising,douglas2007humaine}, emotions were elicited via interaction with a computer avatar. The Geneva multi-modal emotion portrayals (GEMEP) database \cite{glowinski2008technique} compiled more than 7,000 audio-video portrayals of 18 emotions by 10 actors. Similarly, the LIRIS-ACCEDE database \cite{baveye2015liris} focused on upper body emotional gestures from 64 subjects, captured using the Kinect sensor \cite{saha2014study}. Other datasets like emoFBVP \cite{ranganathan2016multimodal} and Emilya \cite{fourati2014emilya} also featured skeletal tracking and 3D body movements of posed emotions.

More recent studies have focused on spontaneous emotional gestures, which present a greater challenge due to their natural and unplanned nature. The Theater dataset \cite{kipp2009gesture} extracted emotional gesture video clips from two movies, providing real-world emotional expressions. Similarly, BoLD (Body Language Dataset) \cite{luo2020arbee} collected in-the-wild perceived emotional data from movies and reality TV shows, segmented based on body movements and annotated with perceived emotional expressions. The Spontaneous Micro-Gesture (SMG) dataset \cite{chen2019analyze} focused on subtle body movements that involuntarily reveal hidden emotions, marking the first dataset of its kind to study MGs under laboratory conditions.

\subsection{Gesture-based emotion analysis}

In comparison to facial expression analysis, studies focusing on body gesture-based emotion recognition are still limited. However, body gestures offer unique advantages as they can reveal hidden emotions, especially when facial expressions are intentionally controlled or masked.

Early work in the field explored multimodal approaches to emotion recognition, where facial expressions and body gestures were treated as distinct modalities and processed simultaneously \cite{gunes2004face}. In subsequent studies, researchers investigated various techniques for combining gestures and facial expressions to improve emotion recognition performance \cite{gunes2005affect}. For example, Zadeh et al. \cite{zadeh2016multimodal} analyzed emotions by focusing on facial gestures such as head nods and shakes, while Castellano et al. \cite{castellano2007recognising} utilized body movement features like amplitude, speed, and fluidity to infer emotional states.

Recent advancements have introduced sentiment analysis techniques that leverage transcription data extracted from videos \cite{stappen2021sentiment}. Another study carried out emotion recognition tasks on skeletal data, utilizing 3D body gestures captured using depth sensors like the Kinect \cite{saha2014study}. These studies lay the groundwork for understanding the correlation between body movements and emotional states.

This work is a subsequent work of our previous conference paper \cite{liu2021imigue}. In \cite{liu2021imigue}, we introduced the preliminary version of our dataset that contains over 300 videos with annotated MGs. In this paper, we scale up the original dataset to iMiGUE-3K, substantially increasing the number of subjects, videos, and MG instances, thereby enabling more robust and generalizable learning in real-world settings. It also provides the benchmark broadened to more downstream tasks beyond only MG recognition. A brief comparison between these two datasets is shown in Table~\ref{tab:comparison_imigue_imigue3k}. Besides, in this paper, we propose a data sampling strategy to generate high-quality clips supporting SSL. Also, we introduce a series of foundation models for transferable MG representation learning, showing that with SSL on large-scale unlabeled data, simple linear probing can already achieve competitive performance against supervised methods among downstream tasks. 
\begin{table}[t]
\centering
\caption{Comparison of the dataset version introduced in this paper and the one released in our preliminary conference paper \cite{liu2021imigue}. F: Face, G: Gesture.}
\label{tab:comparison_imigue_imigue3k}
\resizebox{\linewidth}{!}{
\begin{tabular}{lcccc}
\hline
Version & \#Videos & \#Subjects & Duration & Modalities\\ 
\hline
Preliminary \cite{liu2021imigue} (iMiGUE) & 359 & 72 & 0.5-25.8 Min & G \\
Subsequent (iMiGUE-3K) & 3,484 & 332 & 0.2-41.3 Min & F+G \\
\hline
\end{tabular}
}
\end{table}

\subsection{Computer vision methods for gesture recognition}

Recognizing human gestures and actions from videos is a fundamental task in computer vision research. With the advent of deep learning techniques, significant progress has been made in this area, particularly through the use of convolutional neural networks (CNNs), recurrent neural networks (RNNs) and recent Tranformer based networks. \textbf{CNN-based approaches}: Convolutional neural networks have been widely applied to video-based gesture recognition tasks. Ji et al. \cite{ji20123d} introduced the 3D CNN for action recognition, which processes spatio-temporal information in video sequences. Similarly, Du et al. \cite{du2015hierarchical} proposed C3D, a deep 3D CNN architecture, while Hara et al. \cite{hara2018can} extended this idea by incorporating residual learning into the 3D CNN structure, building on ResNet \cite{he2016deep}. Inspired by the Inception architecture \cite{ioffe2015batch}, Carreira and Zisserman \cite{carreira2017quo} proposed I3D, a 3D network designed to capture detailed action features. Feichtenhofer et al. \cite{feichtenhofer2019slowfast} introduced the SlowFast network, which processes video data at different temporal resolutions to optimize computational efficiency. \textbf{RNN-based approaches}: Recurrent neural networks (RNNs), particularly long short-term memory (LSTM) networks, have also been employed to capture temporal dependencies in gesture recognition tasks, enabling models to learn from sequences of movements over time. \textbf{Transformer-based approaches}: Recently, transformers and attention mechanisms have demonstrated superior performance in both image and video processing tasks \cite{dosovitskiy2020image,arnab2021vivit,liu2022video}. The Video Swin Transformer \cite{liu2022video}, for example, has achieved state-of-the-art results in video-based gesture and action recognition by modeling long-range dependencies and capturing fine-grained details in video data.


\subsection{Foundation Models in Computer Vision}
Foundation models are trained on a large-scale, broad data and can be adapted to a range of downstream tasks \cite{awais2025foundation}. In computer vision, this paradigm has progressed from scalable visual encoders to multimodal models. For backbone design, vision Transformers and hierarchical Transformers provide generic backbones for large-scale transfer \cite{dosovitskiy2020image,liu2021swin}. As for the learning paradigm, contrastive learning and masked autoencoders enable the model to learn reusable visual representations without dense human annotations \cite{he2020momentum,chen2020simple,he2022masked}. Vision-language pre-training has shifted visual foundation models from closed-set recognition toward open-world semantic transfer, where language serves as both scalable supervision and a flexible query interface for high-level visual understanding \cite{radford2021learning,jia2021scaling,wang2023image,zhai2023sigmoid,xiao2024florence,chen2024internvl}. Beyond image-level semantics, promptable perception models extend this paradigm to dense visual prediction by conditioning region- and pixel-level outputs on interactive or textual prompts \cite{li2022grounded,liu2024grounding,ravi2025sam,carion2025sam}. In parallel, generative foundation models learn reusable visual priors that support generation-oriented vision problems \cite{ho2020denoising,rombach2022high}. For dynamic visual data, video foundation models incorporate temporal structure into masked or multimodal pre-training, generalizing foundation-model learning from static appearance representations to spatial-temporal visual understanding \cite{tong2022videomae,wang2023videomae,wang2024internvideo2}.

When the skeleton modality is involved, recent studies have begun to transfer the foundation-model paradigm to structured human-motion data rather than merely designing task-specific networks. One line of work pre-trains motion encoders on heterogeneous pose and motion corpora to learn transferable kinematic priors \cite{zhu2023motionbert}. Another line adopts masked skeleton or motion modeling, where the model learns to recover missing joints, temporal segments, or motion tokens, thereby capturing spatial-temporal dependencies beyond supervised action labels \cite{yan2023skeletonmae,mao2023masked}. More recent efforts further broaden the task and supervision space by introducing multimodal skeleton representation learning, prompt- or in-context-conditioned skeleton understanding, diffusion-based unified motion modeling, and human-centric generalist formulations \cite{sun2023unified,wang2024skeleton,wu2024macdiff,wang2025hulk,usdrl}. 

\section{The iMiGUE-3K dataset}
\label{sec3}
\begin{figure*}[!htp]
    \centering
    \includegraphics[width=0.85\textwidth]{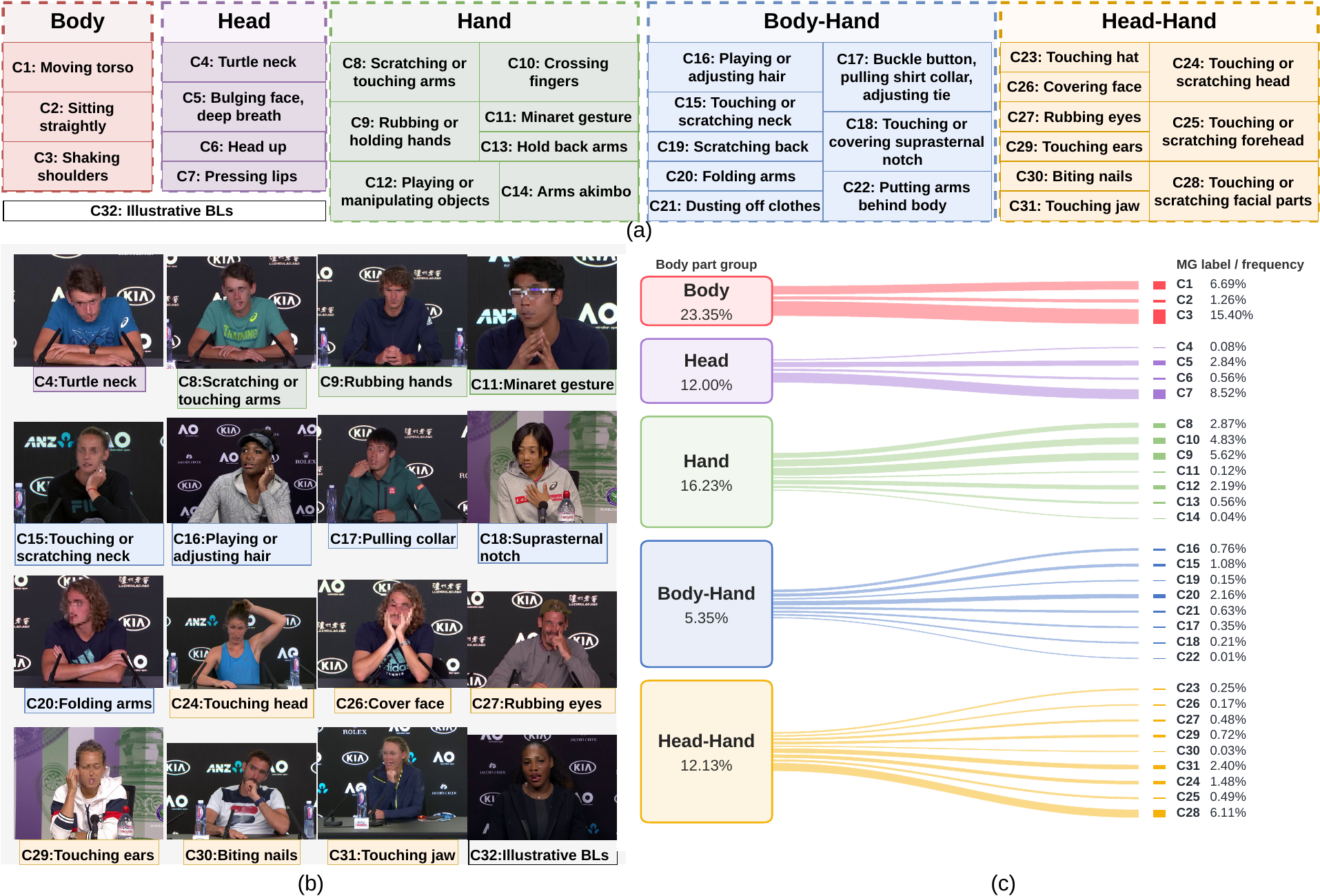}
    \caption{
        Overview of the annotation schema and category distribution in iMiGUE-3K.
        (a) The 32 annotation labels are organized into five body-part-related groups: Body, Head, Hand, Body-Hand, and Head-Hand, where C32 denotes illustrative body languages (illustrative bls).
        (b) Representative annotated examples from different categories, showing the subtle and localized nature of micro-gesture behaviors in post-match interviews.
        (c) Long-tailed distribution of annotated categories in the labeled subset. Colors indicate the corresponding body-part groups.
    }
    \label{fig:mg_overall_1}
\end{figure*}

\begin{figure*}[!htp]
    \centering
    \includegraphics[width=\linewidth]{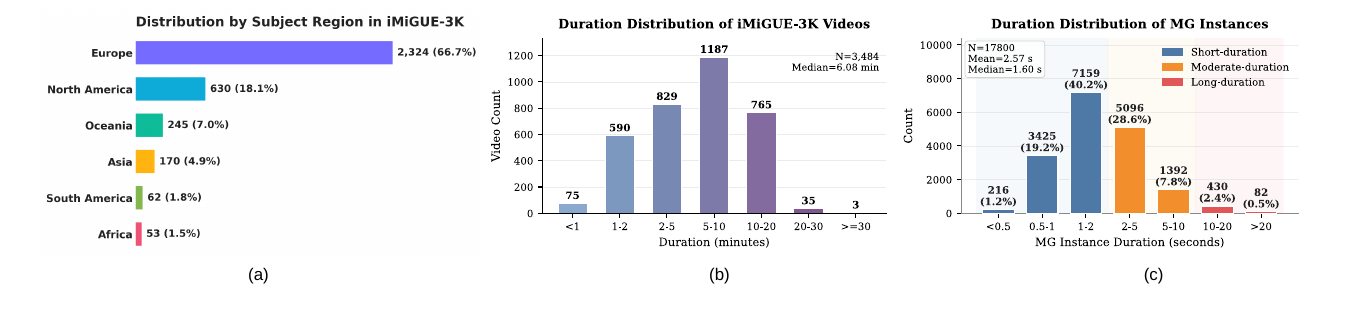}
    \caption{
        Statistics of the iMiGUE-3K dataset.
        (a) Video-Level Distribution by Subject Region in iMiGUE-3K.
        (b) Duration distribution of the videos in iMiGUE-3K.
        (c) Duration distribution of annotated MG instances. Most MG instances are short-lived, while only a small fraction exhibits longer temporal spans, highlighting the temporally subtle and imbalanced nature of spontaneous MG behaviors. Noteably, the class of "illustrative BLs" is not included.
    }
    \label{fig:mg_overall_2}
\end{figure*}

\subsection{Dataset collection}
We constructed the iMiGUE-3K dataset by collecting videos from post-match press conferences of professional athletes. In these scenarios, athletes are interviewed by journalists immediately after intense matches, giving them little time to prepare for responses. This real-time interaction often causes athletes to inadvertently display emotion-related micro-gestures (MGs) despite their attempts to control their verbal responses. As research has shown, “the subconscious mind acts automatically and independently of our verbal statements” \citep{pease2008definitive}, making these post-match interviews ideal for capturing spontaneous emotional cues.

The outcome of the match (win or lose) serves as a natural emotional trigger, with winning players likely experiencing positive emotions and losing players experiencing negative ones. For the purposes of emotion AI research, understanding and detecting these MGs is crucial, as they provide a window into the emotional state of the players that might otherwise remain hidden from verbal communication.

The press conferences from Grand Slam tennis tournaments were chosen as the primary data source for iMiGUE-3K, due to several key advantages:
First, a large number of publicly available high-quality videos (minimum 720p resolution) ensures that even subtle micro-gestures are preserved in detail.
Second, consistent background conditions without interference from the environment—post-match press conferences feature a static advertising wall as the backdrop, which minimizes distractions.
Third, a diverse range of cultures and nationalities is represented by players from all over the world.
Lastly, a well-balanced gender distribution—each Grand Slam tournament features an equal number of male and female players (128 men and 128 women), making it easier to create a balanced dataset in terms of gender representation.

\subsection{Dataset statistics and properties}

iMiGUE-3K is a large-scale in-the-wild dataset consisting of 3,484 video samples collected from public post-match interviews of professional athletes over the past seven years. The dataset covers 332 distinct subjects with diverse cultural and behavioral backgrounds, providing a rich and realistic testbed for studying MG-based emotion understanding, shown in Fig.~\ref{fig:mg_overall_2}(a). All videos are of high quality (typically 1280$\times$720 resolution at 25 FPS), with durations ranging from 0.2 to 41.3 minutes. The duration distribution of the collected videos is shown in Fig.~\ref{fig:mg_overall_2}(b). Table~\ref{tab1} summarizes the main statistics of iMiGUE-3K and compares it with existing gesture-based emotion datasets. Besides, Table~\ref{tab2} provides the statistical properties of iMiGUE-3K.

To support both self-supervised pretraining and supervised downstream evaluation, iMiGUE-3K is partitioned into two subsets: a labeled subset and an unlabeled subset. Since the partition is performed at the video level, the labeled and unlabeled subsets are not necessarily subject-independent. The labeled one modified based on our previous study \cite{liu2021imigue}, namely iMiGUE-3K-labeled, contains 341 videos, where the annotation process is elaborated in Sec.~\ref{sec:dataset_annotation}. The remaining, called iMiGUE-3K-unlabeled, belongs to the unlabeled subset. The statistics of the subset partition are summarized in Table~\ref{tab:imigue3k_subset_statistics}. In this context, the larger-scale unlabeled videos contribute to the pretraining of foundation models via SSL, and the pretrained model can be evaluated using the labeled subset. 

For iMiGUE-3K-labeled, a total of 17,800 MG instances were identified and annotated across 32 different categories. On average, each video contains approximately 52 MG instances, with instance durations ranging from 0.18 to 80.92 seconds (mean duration of 2.55 seconds). 
As shown in Fig.~\ref{fig:mg_overall_2}(c), most MG instances are short-lived, while a small fraction exhibits longer temporal spans. This wide temporal range reflects the inherently spontaneous and unconstrained nature of MGs. 

A key characteristic of iMiGUE-3K is its long-tailed distribution over gesture categories. As shown in Fig.~\ref{fig:mg_overall_1}(c), a small number of frequently occurring gestures dominate the labeled subset, while many categories remain underrepresented. This imbalance naturally arises in real-world spontaneous behaviors \citep{yan2014casme,li2013spontaneous,davison2016samm} and poses significant challenges for conventional supervised learning methods, while also highlighting the importance of transferable and generalizable representation learning.

\begin{table}[ht]
\begin{center}
\centering
\caption{Statistics of the iMiGUE-3K dataset.} 
\begin{tabular}{|l|c|}
  \hline
  Number of videos & 3,484\\
  \hline
  Video duration range & 0.2 $\sim$ 41.3\\
  \hline
  Video resolution & 1280 $\times$ 720\\
  \hline
  Number of gesture types & 32\\
  \hline
  Number of subjects (F/M) & 332 (171/161)\\
  \hline
  Number of matches (W/L) & 3,484 (2791/693)\\
\hline
\end{tabular}
\label{tab2}
\end{center}
\end{table}


\begin{table}[t]
\centering
\begin{threeparttable}
\caption{Statistics of the partitioned subsets of iMiGUE-3K. 
\textit{Subset-l} denotes iMiGUE-3K-labeled, and \textit{Subset-u} denotes iMiGUE-3K-unlabeled.}
\label{tab:imigue3k_subset_statistics}
\begin{tabular}{lcccc}
\toprule
Subset & \#Videos & \#Subjects & \#MG Instances & \#Win/\#Lose \\ 
\midrule
\textit{Subset-l} & 341   & 75  & 17,800              & 251/90 \\
\textit{Subset-u} & 3,143 & 319 & 120K\tnote{*}       & 2,540/603 \\
\bottomrule
\end{tabular}
\begin{tablenotes}[flushleft]
\footnotesize
\item[*] Sampled clips for self-supervised learning, not exact MG instances. More details are provided in Sec.~\ref{sec:sample_strategy}.
\end{tablenotes}
\end{threeparttable}
\end{table}

\subsection{Dataset annotation}
\label{sec:dataset_annotation}

The annotation of iMiGUE-3K contains two levels: 
(1) the temporal localization and category annotation of MG instances, and 
(2) the video-level emotion annotation. 
Due to the large scale of iMiGUE-3K, we do not annotate MG instances for the entire dataset. Instead, we provide a human-annotated labeled subset, iMiGUE-3K-labeled, while the remaining videos are used as unlabeled data for self-supervised pretraining.

\textbf{MG Labeling.}
For instance-level MG annotation, annotators were required to spot the start and end points of each MG instance and assign it to one of the predefined categories. Based on psychological studies of nonverbal behaviors~\citep{ekman2009telling,pease2008definitive,navarro2008every}, the MG taxonomy is organized into five body-part-related groups: Head, Body, Hand, Body-Hand, and Head-Hand. These groups contain 31 MG categories, together with one non-MG category for illustrative gestures, resulting in 32 annotation labels in total, as shown in Fig.~\ref{fig:mg_overall_1}.

Five human annotators participated in the annotation process. Before formal annotation, all annotators underwent training sessions to unify their understanding of the MG taxonomy, category definitions, and temporal boundary criteria. This session guarantees the consistency of labeling across different annotators. The training process included reviewing category descriptions, inspecting representative video examples, and discussing ambiguous cases, especially for subtle or short-duration MGs.

To ensure annotation quality, each video in iMiGUE-3K-labeled was independently annotated by two annotators. Their annotations were then summarized and cross-checked. Inconsistent cases, including category disagreements and temporal boundary mismatches, were discussed to reach a consensus. Cases that remained ambiguous after discussion were excluded from the final labeled subset. The annotation reliability was measured by the agreement ratio:
\begin{equation}
R(Anno_i,Anno_j)=
\frac{2 \times MG(Anno_i,Anno_j)}
{\#MG(Anno_i)+\#MG(Anno_j)},
\label{imgeq1}
\end{equation}
where $MG(Anno_i,Anno_j)$ denotes the number of matched MG instances between annotators $Anno_i$ and $Anno_j$, and $\#MG(Anno_i)$ and $\#MG(Anno_j)$ denote the numbers of MG instances annotated by the two annotators, respectively. The average inter-annotator agreement was $R_{\mathrm{avg}}=0.81$, indicating high annotation consistency.

\textbf{Video-Level Emotion Annotation.}
In addition to instance-level MG annotations, each video is assigned a video-level emotion label according to the match outcome. Specifically, interviews of winning players are associated with positive affective contexts, while interviews of losing players are associated with negative affective contexts. These labels are derived from objective match results rather than subjective emotion judgments, and are used as video-level supervision for MG-based emotion-related tasks.

To obtain the skeleton information, for each source video $v$, RTMPose~\cite{jiang2023rtmpose} is applied frame by frame to estimate a whole-body pose sequence
\begin{equation}
\mathbf{S}^{(v)} \in \mathbb{R}^{C \times T_v \times J \times M},
\end{equation}
where $C=3$ denotes the $(x,y,c)$ channels, $T_v$ is the number of frames of video $v$, $J$ is the number of body joints, and $M=1$ denotes the primary interviewee in the video.

\subsection{Experiment protocols}
To evaluate foundation models on iMiGUE-3K, we define five downstream tasks: unsupervised MG recognition, Semi-supervised MG recognition, supervised MG recognition, MG retrieval, and MG-based emotion recognition. For each downstream task, the foundation models are first pretrained on iMiGUE-3K-unlabeled and then evaluated on iMiGUE-3K-labeled under task-specific protocols.

For downstream tasks, the labeled subset is split into training and testing sets with a subject separation ratio of approximately 1:1. In total, 37 subjects were used for training and 38 for testing, resulting in 13,478 and 4,322 MGs in the training and testing sets, respectively, as shown in Table~\ref{tab:imigue3k_labeled}.

\begin{table}[t]
\centering
\caption{Statistics of the iMiGUE-3K-labeled dataset.}
\label{tab:imigue3k_labeled}
\resizebox{\linewidth}{!}{
\begin{tabular}{lcccc}
\hline
Split & \#Videos & \#Subjects & \#MG Instances & \#Win/Lose\\ 
\hline
Training Set & 248 & 37 & 13,478 & 199/49 \\
Test Set & 93 & 38 & 4,322 & 52/41 \\
\hline
\end{tabular}
}
\end{table}

\section{MG-FMs: A Series of Foundation-Models Pretraining for Transferable Micro-Gesture Representation Learning}
\label{sec4}
MG-FMs, a series of foundation models, are designed to learn transferable micro-gesture representations from large-scale, unlabeled clips. It contains two modality-specific branches: a skeleton branch for modeling compact human-motion dynamics and an RGB branch for modeling appearance and contextual visual cues. Both branches are pretrained on unlabeled clips sampled from iMiGUE-3K and are evaluated under linear-probe and fine-tuning protocols on the labeled subset.
\subsection{Task Definition}

Let $\mathbf{V}\in\mathbb{R}^{T\times H\times W\times 3}$ denote an RGB clip and
$\mathbf{S}\in\mathbb{R}^{C\times T\times J\times M}$ denote its corresponding skeleton sequence,
where $C=3$ corresponds to the $(x,y,c)$ channels, $J$ is the number of joints, and $M$ is the number of persons.
The goal of MG representation learning is to learn modality-specific encoders
$f_{\mathrm{rgb}}(\cdot)$ and $f_{\mathrm{skel}}(\cdot)$ that map the input into compact embeddings:
\begin{equation}
\mathbf{h}_{\mathrm{rgb}} = f_{\mathrm{rgb}}(\mathbf{V}),\qquad
\mathbf{h}_{\mathrm{skel}} = f_{\mathrm{skel}}(\mathbf{S}),
\end{equation}
such that the embeddings capture transferable appearance, motion, and kinematic patterns for downstream MG understanding.

Unlike conventional supervised MG classification, which relies on predefined labels and is sensitive to long-tailed data distributions, we focus on self-supervised representation learning to capture intrinsic motion patterns and improve generalization under limited supervision.

To evaluate representation quality, we adopt the linear probe and fine-tuning protocols, where the model is first pre-trained on the iMiGUE-3K-unlabeled corpus and then fine-tuned on the training split of the iMiGUE-3K-labeled dataset.

\subsection{Pre-training Clip Sampling and Preprocessing}
\label{sec:sample_strategy}
Raw interview videos do not provide temporal MG boundaries, and the labeled MG categories follow a long-tailed distribution. Therefore, assigning pseudo-labels to unlabeled videos would introduce substantial noise, especially for rare categories. We instead construct unlabeled MG-length clips and use them for self-supervised pretraining.
      
To make the sampled clips match the realistic temporal distribution of MGs, we estimate an empirical duration prior from the training set in our previous study \cite{liu2021imigue}. We estimate a duration prior from the valid MG instances in the labeled training set. Let $\mathcal{D}_{\mathrm{dur}}=\{d_n\}_{n=1}^{N}$ denote the set of annotated gesture durations. The empirical duration distribution is defined as
\begin{equation}
p_{\mathrm{dur}}(d)=\frac{1}{N}\sum_{n=1}^{N}\delta(d-d_n),
\end{equation}
where $\delta(\cdot)$ is the Dirac delta function. For each source video, a target duration is sampled as
\begin{equation}
d \sim p_{\mathrm{dur}}(d).
\end{equation}
Given the frame rate $fr_v$, we then uniformly sample a start time
\begin{equation}
s \sim \mathcal{U}\left(0,\frac{T_v}{fr_v}-d\right),
\end{equation}
and convert the temporal interval into frame indices
\begin{equation}
\tau_s = \left\lfloor s fr_v \right\rfloor + 1,\qquad
\tau_e = \left\lceil (s+d) fr_v \right\rceil.
\end{equation}
The corresponding candidate video clip is
\begin{equation}
\mathbf{V}_{\mathrm{cand}}
=
\mathbf{V}^{(v)}_{\tau_s:\tau_e},
\qquad
\Omega=[\tau_s,\tau_e].
\end{equation}

However, directly sampled intervals may still contain severe noise caused by abrupt shot transitions, interviewer cut-ins, replay segments, and unstable pose estimation. To suppress such noise, we impose a visibility constraint on a small set of facial keypoints. Let $\mathcal{A}$ denote the anchor facial keypoint set consisting of the nose, left eye, right eye, left ear, and right ear. A candidate interval is accepted only if all anchors remain confidently detectable throughout the entire clip:
\begin{equation}
\phi(\mathbf{S}_{\mathrm{cand}})
=
\prod_{t=\tau_s}^{\tau_e}
\prod_{j\in\mathcal{A}}
\mathbb{I}\!\left(
\mathbf{S}^{(v)}_{3,t,j,1}>\gamma
\right),
\end{equation}
where $\gamma$ is the confidence threshold for facial landmarks, $\mathbb{I}(\cdot)$ is the indicator function, and $\mathbf{S}^{(v)}$ is the skeleton sequence corresponding to source video $v$.

Here, $\mathbf{S}^{(v)}_{3,t,j,1}$ denotes the confidence channel of keypoint $j$ at frame $t$. A candidate interval is accepted iff $\phi(\mathbf{S}_{\mathrm{cand}})=1$.

To reduce redundancy, accepted clips from the same source video are required to be temporally non-overlapping. Suppose that $\Omega_a=[\tau_s^{(a)},\tau_e^{(a)}]$ and $\Omega_b=[\tau_s^{(b)},\tau_e^{(b)}]$ are two sampled intervals from the same video. They are both retained only if
\begin{equation}
\Omega_a \cap \Omega_b = \emptyset.
\end{equation}
Accordingly, the final sampling process can be formulated as rejection sampling:
\begin{equation}
\mathbf{S}_i =
\mathbf{S}^{(v)}_{:,\tau_s^{(i)}:\tau_e^{(i)},:,:},
\quad
\text{s.t.}\quad
\phi(\mathbf{S}_i)=1,
\quad
\Omega_i \cap \Omega_k = \emptyset,\ \forall\, k<i.
\end{equation}
For each source video, we repeatedly sample candidate intervals until either the predefined quota is reached or the maximum retry budget is exhausted.

\begin{figure}[!htp]
    \centering
    \includegraphics[width=\linewidth]{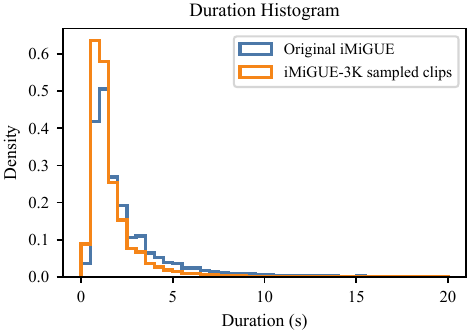}
    \caption{Comparison between duration distributions of iMiGUE \cite{liu2021imigue} and iMiGUE-3K-unlabeled sampled clips. The iMiGUE distribution is the prior used in the sampling process.}
    \label{fig:duration_comparison}
\end{figure}

Using this procedure, we extract at most 50 skeleton clips from each source video. This quota is chosen because it is close to the average number of MG instances per video in iMiGUE-3K-labeled. Because some videos do not contain enough valid intervals under the above constraints, the actual number of clips is smaller than the nominal upper bound. In total, this process produces 120K unlabeled skeleton clips from approximately 3,000 interview videos. These clips are then used as the pre-training corpus, and each sampled clip is represented in the same tensor form as the downstream model input:
\begin{equation}
\mathbf{S}\in\mathbb{R}^{C\times T\times J\times M},
\end{equation}
thereby ensuring consistency between data preprocessing and skeleton-based foundation model training. The comparison between duration distributions of the training set in iMiGUE \cite{liu2021imigue} and the sampled clips is shown in Fig.~\ref{fig:duration_comparison}. For readability, only durations shorter than 20 seconds are displayed.

We also explored the influence of the pre-defined number of skeleton clips. In a very pilot study, we conduct a simple ablation study on the test set of iMiGUE: a similar random sampling process is implemented by changing only the facial-landmark confidence threshold, such that the number of generated clips is comparable to the number of ground-truth MG instances in the test set. We use the intersection over detection (IoD) to evaluate the MG occupancy in different sets of randomly-sampled clips \cite{lu2021weakly,xu2024efficient}. In brief, the MG-IoD precision is expressed as:
\begin{equation}
\begin{aligned}
    & IoD=\frac{\ell(\Omega \cap \mathcal{G})}{\ell(\Omega)}, \\
    & Prec_{MG}^{IoD@0.5}=\frac{1}{|\mathcal{W}_{rand}|}\sum_{\Omega \in \mathcal{W}_{rand}}\mathbb{I}[IoD(\Omega) \geq 0.5],
\end{aligned}
\end{equation}
where $\mathcal{W}_{rand}$ represents the set of randomly-sampled clips, and $\mathcal{G}$ denotes the union of groundtruth MG intervals. Table~\ref{tab:pilot_study_random_sampling_ablation} and Table~\ref{tab:pilot_study_confidence_threshold_ablation} shows that the MG occupancy remains relatively stable under different sampling hyperparameters.
show

\begin{table}[h]
\centering
\caption{Effect of the predefined sampling quota. The facial-landmark confidence threshold is fixed at 0.3.}
\begin{tabular}{lcc}
\hline
Sampling Number & Generated Samples & MG-IoD Precision \\
\hline
50 & 4374 & 39.39\% \\
80 & 6159 & 40.57\% \\
100 & 6919 & 40.16\% \\
\hline
\end{tabular}
\label{tab:pilot_study_random_sampling_ablation}
\end{table}

\begin{table}[h]
\centering
\caption{Effect of the facial-landmark confidence threshold. The predefined sampling quota is fixed at 80.}
\begin{tabular}{lcc}
\hline
Confidence Threshold & Generated Samples & MG-IoD Precision \\
\hline
0.1 & 6726 & 39.47\% \\
0.3 & 6159 & 40.57\% \\
0.5 & 5708 & 39.87\% \\
0.75 & 4324 & 40.24\% \\
\hline
\end{tabular}
\label{tab:pilot_study_confidence_threshold_ablation}
\end{table}

Based on the skeleton sequences extracted in Sec.~\ref{sec:dataset_annotation}, we construct the input sequences for the skeleton foundation model. First, a compact upper-body-centric skeleton representation is constructed from the whole-body pose annotations, which preserves body, hand, and selected facial landmarks, because post-match interviews primarily expose the upper body and MGs are mainly expressed through upper-body movements. Second, the coordinates of skeleton sequences are normalized to a canonical image scale and centered with respect to the shoulder reference, yielding a person-centric representation. Third, for the skeleton foundation model, variable-length clips are padded or truncated to a fixed temporal length while retaining the number of valid frames. 

For the RGB branch, clips are cropped from the same temporal intervals as the skeleton clips, ensuring temporal alignment between the two modalities.

\subsection{MG-FMs in Skeleton Modality}
\label{sec:skeleton}
\subsubsection{MG-FM-Skele}

For the skeleton modality, we instantiate MG-FM-Skele by adapting USDRL~\cite{usdrl} to large-scale 2D skeleton pre-training on iMiGUE-3K. We retain its dual-stream Dense Spatio-Temporal Encoder (DSTE) and multi-grained feature decorrelation objective, while redesigning the input formulation, augmentation pipeline, and scaling strategy for our large-scale skeleton dataset.

Given a skeleton clip $\mathbf{S}\in\mathbb{R}^{C\times T\times J\times M}$, we rearrange it into a temporal token sequence and a spatial token sequence:
\begin{equation}
\mathbf{S}^{t}\in\mathbb{R}^{T\times MJC},\qquad
\mathbf{S}^{s}\in\mathbb{R}^{MJ\times TC}.
\end{equation}
These two token sequences are fed into the temporal and spatial streams of DSTE to model motion dynamics and joint structure in parallel. The resulting temporal and spatial representations are further aggregated and mapped by temporal, spatial, and instance projectors to produce three complementary embeddings for self-supervised pre-training.

During pre-training, each skeleton clip is transformed into four augmented views, denoted by $\{\tilde{\mathbf{S}}^{(m)}\}_{m=1}^{4}$. Beyond the original crop-resize and spatial perturbations in USDRL, we adopt stronger 2D skeleton augmentations tailored to iMiGUE-3K, including speed jitter, frame dropping, confidence-aware joint corruption, pose augmentation, and random 2D rigid transformations. After dual-stream encoding and projection, the $m$-th view produces temporal, spatial, and instance embeddings
\begin{equation}
\mathbf{h}_{m}^{t},\ \mathbf{h}_{m}^{s},\ \mathbf{h}_{m}^{i}\in\mathbb{R}^{D},\qquad m=1,\ldots,4.
\end{equation}

Let $\mathbf{H}^{r}=\{\mathbf{h}_{m}^{r}\}_{m=1}^{4}$ denote the four-view embeddings in domain $r\in\{t,s,i\}$. We first define a domain-wise feature decorrelation loss
\begin{equation}
\mathcal{L}_{\mathrm{fd}}(\mathbf{H}^{r})
=
\lambda_{\mathrm{sim}}\mathcal{L}_{\mathrm{sim}}(\mathbf{H}^{r})
+
\lambda_{\mathrm{vac}}\mathcal{L}_{\mathrm{vac}}(\mathbf{H}^{r})
+
\lambda_{\mathrm{xcorr}}\mathcal{L}_{\mathrm{xcorr}}(\mathbf{H}^{r}),
\end{equation}
where $\lambda_{\mathrm{sim}}=5$, $\lambda_{\mathrm{vac}}=1$, and $\lambda_{\mathrm{xcorr}}=10^{-3}$.
Following USDRL, the final pre-training objective is
\begin{equation}
\mathcal{L}_{\mathrm{skel}}
=
\mathcal{L}_{\mathrm{fd}}(\mathbf{H}^{i})
+
\tau\left(
\mathcal{L}_{\mathrm{fd}}(\mathbf{H}^{t})
+
\mathcal{L}_{\mathrm{fd}}(\mathbf{H}^{s})
\right),
\qquad \tau=0.5.
\end{equation}

Thus, MG-FM-Skele preserves the feature-decorrelation principle of USDRL while being adapted to large-scale 2D skeleton foundation pre-training on iMiGUE-3K.

\subsection{MG-FMs in RGB Modality}
\label{sec:rgb}
\subsubsection{MG-FM-RGB}

MG-FM-RGB serves as the RGB branch of MG-FMs. It adopts a SMILE-style teacher-student masked video modeling framework to perform semantic pretraining on large-scale unlabeled RGB clips from iMiGUE-3K \cite{smile}. Given an RGB clip $\mathbf{V}\in\mathbb{R}^{T\times H\times W\times 3}$, we partition it into $N$ non-overlapping spatio-temporal tubelets:
\begin{equation}
\mathcal{T}(\mathbf{V})=\{\mathbf{t}_i\}_{i=1}^{N},
\qquad
N=\frac{T}{\tau}\cdot\frac{H}{P}\cdot\frac{W}{P},
\end{equation}
where $(\tau,P,P)=(2,16,16)$ in our implementation. A subset $\mathcal{M}\subset\{1,\ldots,N\}$ is masked and the visible tubelets are fed into a VideoMAE-style student transformer. Instead of reconstructing raw pixels, MG-FM-RGB reconstructs high-level semantic targets extracted by a frozen ViCLIP visual encoder. For each masked tubelet $i\in\mathcal{M}$, let $\mathbf{q}_i\in\mathbb{R}^{d}$ denote the teacher target feature extracted from the first frame associated with that tubelet, and let $\hat{\mathbf{q}}_i$ denote the corresponding student prediction. The RGB pre-training objective is defined as
\begin{equation}
\mathcal{L}_{\mathrm{rgb}}
=
\frac{1}{|\mathcal{M}|}
\sum_{i\in\mathcal{M}}
\left\|
\hat{\mathbf{q}}_i - \mathbf{q}_i
\right\|_2^2.
\label{eq:smile_loss}
\end{equation}

Following SMILE~\cite{smile}, MG-FM-RGB adopts a progressive two-stage pre-training strategy. In Stage I, the synthetic motion augmentation is applied to the input clips, and the semantic masked video modeling with trajectory-aware masking is performed, encouraging the model to focus on motion-sensitive regions. In Stage II, we initialize the model from the Stage-I checkpoint and continue pre-training on the original RGB clips using standard tube masking. The same semantic reconstruction objective in Eq.~\ref{eq:smile_loss} is used in both stages. In this way, MG-FM-RGB combines ViCLIP-based semantic supervision, motion-aware masking, and large-scale RGB pre-training on iMiGUE-3K.
\section{Unsupervised MG Recognition}
\label{sec:unsupervised_mg_recognition}
This task evaluates the transferability of the pretrained MG representations on iMiGUE-3K-labeled. Given a temporally localized clip containing an MG instance, the model predicts its MG category. For MG-FM-Skele and MG-FM-RGB, the pretrained encoders are frozen, and only a linear classifier is trained on the labeled training split. Therefore, this protocol evaluates the quality of the learned representations rather than end-to-end supervised recognition performance.

For skeleton-based recognition, we evaluate joint, bone, and motion streams, as well as their multi-stream combination. The unsupervised algorithms in \cite{liu2021imigue} are compared. The RGB branch is evaluated under the same linear-probe protocol.

Table~\ref{tab:linear_probe} shows the linear probing performance on the iMiGUE-3K test set. Our self-supervised MG-FM achieves competitive performance compared to supervised methods. Notably, the skeleton-based multi-stream variant (joint + bone + motion) achieves \textbf{57.61\% Top-1 accuracy}, which is on par with the supervised 4-stream BlockGCN (57.85\%), demonstrating strong transferability of the learned representations. These results indicate that MG-FM-Skele effectively captures fine-grained motion patterns and generalizes well across gesture categories, even without explicit supervision. Besides, MG-FM-RGB can only get 43.59\% Top-1 accuracy, implying that learning transferable representations from raw RGB data remains more challenging due to background noise and irrelevant visual cues.

\begin{table}[h]
\centering
\caption{Results of unsupervised MG recognition on iMiGUE-3K-labeled. Notably, these compared methods are skeleton-based and trained supervisely. The best results are marked in bold, and the second-best results are underlined. The same notation is used below.}
\begin{tabular}{lcc}
\hline
Method & Top-1 (\%) & Top-5 (\%) \\
\hline
P\&C \cite{su2020predict}& 31.67 & 64.93\\
U-S-VAE Z \cite{liu2021imigue} & 32.43 & 64.30\\
\hline
Ours \\
\hline
MG-FM-RGB (RGB) & 43.59 & 80.48 \\
MG-FM-Skele (joint) & \underline{53.65} & 83.76 \\
MG-FM-Skele (bone) & 52.34 &  \underline{83.99} \\
MG-FM-Skele (motion) & 51.94 & 83.39 \\
MG-FM-Skele (3-stream) & \textbf{57.61} & \textbf{86.97} \\
\hline
\end{tabular}
\label{tab:linear_probe}
\end{table}

\section{Semi-Supervised MG Recognition}
\label{sec:semisupervised_mg_recognition}
In the semi-supervised setting, the pretrained models are fine-tuned using only 10\% of the labeled training samples randomly selected from iMiGUE-3K-labeled. Because MG categories follow a long-tailed distribution, this subset may not cover all categories, making the setting particularly challenging. Results are shown in Table~\ref{tab:semi_supervised}. This shows that MG-FM-RGB improves substantially from its linear-probe performance, reaching 60.16\% Top-1 accuracy with only 10\% labeled data. This suggests that RGB representations benefit strongly from limited task-specific adaptation. The skeleton branch remains usable under sparse supervision, but its performance is lower than the linear-probe multi-stream setting, indicating that fine-tuning with very limited and long-tailed labels may not fully preserve the pretrained skeleton representation.
\begin{table}[h]
\centering
\caption{Results of semi-supervised MG recognition on iMiGUE-3K-labeled using 10\% of the labeled training samples for fine-tuning.}
\begin{tabular}{lcc}
\hline
Method & Top-1 (\%) & Top-5 (\%) \\
\hline
MG-FM-Skele (joint) & 44.73 &  \underline{81.16} \\
MG-FM-Skele (bone) & 44.19  & 79.67 \\
MG-FM-Skele (motion) &  \underline{46.89} & 77.14 \\
MG-FM-RGB & \textbf{60.16} & \textbf{89.97} \\
\hline
\end{tabular}
\label{tab:semi_supervised}
\end{table}

\section{Supervised MG Recognition}
\label{sec:supervised_mg_recognition}
For supervised MG recognition, the pre-trained models are fine-tuned using all samples in the training set of iMiGUE-3K-labeled. 

Table~\ref{tab:fine_tune} compares MG-FM with prior RGB-based state-of-the-art methods under the full fine-tuning setting. MG-FM-RGB achieves the highest accuracy among the compared methods, indicating that RGB pretraining provides strong initialization for fully supervised adaptation. Besides, it reveals that fine-tuning does not consistently improve MG-FM-Skele over its linear-probe performance, suggesting that the representations of the pre-trained model are highly separable. Such an observation is consistent with prior findings that probing can be highly effective when the source and target domains are relatively similar \cite{wu2024structured}. Meanwhile, full fine-tuning may only provide marginal gains when the pretrained features are already strong, since updating the backbone can distort the pretrained representation space \cite{kumar2022fine}.  

\begin{table}[h]
\centering
\caption{Results of supervised MG recognition on iMiGUE-3K-labeled.}
\begin{tabular}{lcc}
\hline
Method & Top-1 (\%) & Top-5 (\%) \\
\hline
RGB-based Methods \\
\hline
Temporal Relational + RGB (TRN) \cite{trn} &  55.24 & - \\
Temporal Shift + RGB (TSM) \cite{tsm} & 58.77 & - \\
CLIP-MG \cite{clip-mg} & 61.82 & - \\
MMGesture (RGB-only) \cite{mmgesture} &  \underline{66.62} & - \\
MG-FM-RGB (\textbf{Ours}) & \textbf{69.42} & \textbf{93.08} \\
\hline
Skeleton-based Method \\
\hline
BlockGCN (joint) \cite{blockgcn} & 51.94 & 82.96 \\
BlockGCN (bone) & 52.09 & 82.74 \\
BlockGCN (motion) & 52.18 & 81.81 \\
BlockGCN (4-stream) & \textbf{57.85} &  \underline{86.51} \\
PoseC3D (joint) \cite{posec3d} & 56.13 & 85.40 \\
MG-FM-Skele (joint) (\textbf{Ours}) & 53.16 & 85.45 \\
MG-FM-Skele (bone) (\textbf{Ours})& 52.73 & 85.14 \\
MG-FM-Skele (motion) (\textbf{Ours})& 51.62 & 82.76 \\
MG-FM-Skele (3-stream)  (\textbf{Ours}) & \ \underline{57.52} & \textbf{86.78} \\
\hline
\end{tabular}
\label{tab:fine_tune}
\end{table}

Taken together, the results reveal a modality-dependent behavior. MG-FM-Skele produces skeleton representations that are more linearly separable after self-supervised pretraining, whereas MG-FM-RGB benefits more from task-specific fine-tuning and ultimately achieves stronger supervised performance. This suggests that skeleton inputs provide compact kinematic cues that align well with MG categories, while RGB inputs contain richer appearance and contextual information that requires supervised adaptation to be effectively exploited.

Moreover, a multimodal ensemble combining the skeleton-based and RGB-based models is evaluated. Specifically, the final prediction is obtained by weighted aggregation of the prediction scores from the four streams:
\begin{equation}
    P_{\mathrm{ens}}=\sum_{m\in\mathcal{M}} w_m P_m,
    \quad
    \mathcal{M}=\{\mathrm{joint}, \mathrm{bone}, \mathrm{motion}, \mathrm{RGB}\}.
\end{equation}
The comparison with prior supervised ensemble methods is reported in Table~\ref{tab:ensemble}, where fine-tuning models are leveraged. Under the fair setting where both skeleton and RGB modalities are used, MG-FM achieves competitive performance and obtains the best overall results, with 71.60 Top-1 and 95.07 Top-5 accuracy. Despite relying on pre-training with unlabeled data, our method still matches or slightly surpasses existing supervised methods, indicating that large-scale unlabeled pre-training can produce transferable representations that are highly effective for downstream micro-gesture recognition.

\begin{table}[h]
\centering
\caption{Multimodal ensemble results on iMiGUE-3K-labeled. All compared methods use both skeleton and RGB modalities. }
\begin{tabular}{lcc}
\hline
Method & Top-1 (\%) & Top-5 (\%) \\
\hline
CLIP + 3DCNN \cite{wang2024multimodal} & 68.90 & 92.43 \\
Res2Net3D \cite{huang2024multi} & 70.19 &  \underline{93.69}\\
Prototype Learning \cite{chen2024prototype}  & 70.25 & - \\
MM-Gesture \cite{mmgesture} &  \underline{71.42} & - \\
MG-FM (\textbf{Ours}) & \textbf{71.60} & \textbf{95.07} \\
\hline
\end{tabular}
\label{tab:ensemble}
\end{table}

\section{Training-Free MG Retrieval}
\label{sec:mg_retrieval}
This task evaluates the pretrained encoders under a training-free retrieval protocol. Following common action retrieval settings, the compact embeddings $\mathbf{h}_{skel}$ and $\mathbf{h}_{rgb}$ are extracted and directly applied to the nearest-neighbor search, obviating the need for any additional training. Specifically, the labeled training MG instances are used to construct a gallery representation bank, while each test MG instance is treated as a query. Given a query embedding, the cosine similarities with all gallery embeddings are calculated and ranked. 

Table~\ref{tab:retrieval_skeleton} reports the MG retrieval results on iMiGUE-3K-labeled. It shows that MG-FM-Skele consistently outperforms MG-FM-RGB. This suggests that when only pre-training is taken, the skeleton-based model learns more separable representations than the RGB-based model, which is aligned with Sec.~\ref{sec:unsupervised_mg_recognition}.

\begin{table}[h]
\centering
\caption{Results of training-free MG retrieval on iMiGUE-3K-labeled.}
\begin{tabular}{lcc}
\hline
Method & Top-1 (\%) & Top-5 (\%) \\
\hline
MG-FM-RGB & 22.56 & 53.06 \\
MG-FM-Skele (joint) & 37.12 & \textbf{67.04}\\
MG-FM-Skele (bone) &  \underline{38.46} &  \underline{66.88}\\
MG-FM-Skele (motion) & \textbf{40.06} & 66.86\\
\hline
\end{tabular}
\label{tab:retrieval_skeleton}
\end{table}


\section{MG-based Emotion Recognition}
\label{sec6}

The MG-based emotion recognition task on iMiGUE-3K is formulated as video-level binary classification: given a post-match interview video $v$ of a professional tennis player, the model predicts whether the player won or lost the match, where the outcome serves as a proxy for the player's underlying emotional state (positive or negative). Formally, let $v = \{c_1, c_2, \ldots, c_{N_v}\}$ denote a video decomposed into $N_v$ short clips and let $y \in \{0, 1\}$ denote the corresponding binary label. The objective is to learn a mapping $f : v \rightarrow y$ that aggregates evidence accumulated across clips into a single video-level decision. Compared to the clip-level recognition task, this setting is substantially harder: the supervisory signal is weak (one label per video rather than per clip).

We address this task with a two-stream architecture that explicitly decouples representation learning from temporal aggregation, so that the limited video-level supervision is used only to fit a lightweight aggregator over strong pretrained features. Two frozen foundation models pretrained on iMiGUE-3K-unlabeled serve as clip-level feature extractors (see Sec.~\ref{sec:skeleton} and Sec.~\ref{sec:rgb}). Both encoders remain frozen throughout. For each video, the per-clip features are grouped by video identifier and padded to a fixed sequence length of $N=300$ clips, accompanied by a binary attention mask that indicates valid positions.

Table~\ref{tab:winlose} reports the comparison with two external baselines submitted to 3rd MiGA challenge \cite{Shahane,Haozhebu}. Notably, we compare only with methods that do not use textual inputs, since textual information may reveal the outcome of a tennis match and thus introduce information leakage \cite{kakouros2025sounding}.
ISPCAST uses a transformer-based fusion framework for video and facial features, reaching $0.63$. Haozhebu adopts a dual-stream visual architecture with emotion-aware facial analysis and also achieves $0.63$.
Our two-stream method achieves $0.64$ test accuracy using only two frozen foundation models pretrained on iMiGUE-3K-unlabeled and a single linear aggregator, without additional facial-expression encoders, vision-language models, or per-frame emotion classifiers. The fact that this minimal configuration is competitive with substantially more complex pipelines that explicitly incorporate facial cues underscores two findings: first, that the in-domain RGB foundation model pretrained on iMiGUE-3K already encodes a large fraction of the discriminative information needed for outcome-level emotion inference; and second, that under the present scale of video-level supervision the bottleneck lies in aggregation and class imbalance rather than in clip-level representation quality. 

\begin{table}[t]
\centering
\caption{Results of emotion recognition on iMiGUE-3K-labeled. Accuracy is reported on the held-out test split. Best: \textbf{bold}.}
\label{tab:winlose}
\resizebox{\linewidth}{!}{
\begin{tabular}{llc}
\toprule
Method & Model+Modality & Accuracy \\
\midrule
Method 1 (ISPCAST) \cite{Shahane}    & Transformer+RGB+Face      & 0.63 \\
Method 2 (Haozhebu) \cite{Haozhebu}  & CNN+RGB+Face              & 0.63 \\
Ours                  & Transformer+RGB+Skel              & \textbf{0.64} \\
\bottomrule
\end{tabular}
}
\end{table}

\section{Conclusion}
\label{sec8}
In this paper, we proposed a new large-scale dataset for micro-gestures (MGs) analysis, consisting of more than 3k videos from post-match press conferences of professional athletes. This dataset, combining the labeled subset and the unlabeled subset, offers the largest-scale dataset of MGs and thus enables more robust and generalizable learning in real-world settings.

Based on this dataset, we introduced the self-supervised learning (SSL) paradigm into MG analysis, thereby broadening the scope of the benchmark beyond supervised MG recognition to more tasks, including unsupervised MG recognition, semi-supervised MG recognition, MG retrieval, and MG emotion recognition. Furthermore, we propose a data sampling strategy to generate high-quality clips for SSL from unlabeled videos. Moreover, we introduced MG-FM, a series of foundation models for transferable MG representation learning. Results of extensive experiments clearly demonstrate the superiority of pre-training on a massive dataset: our methods outperform prior supervised methods. The iMiGUE-3K dataset, combined with the proposed MG-FM, sets a new benchmark for future research in MG-based analysis.

In conclusion, this work makes significant contributions to the field of AI applications by advancing the analysis of subtle, involuntary MGs, which is important for emotion AI. The insights and methodologies presented here have broad implications for applications in human-computer interaction, affective computing, and psychological diagnostics, paving the way for future innovations in emotion-aware systems.

\ifCLASSOPTIONcaptionsoff
  \newpage
\fi

\bibliographystyle{IEEEtran}
\bibliography{refs}




\end{document}